%% file: egpaper_final.tex
\documentclass[10pt,twocolumn,letterpaper]{article}

\usepackage{cvpr}
\usepackage{times}
\usepackage{epsfig}
\usepackage{graphicx}
\usepackage{amsmath}
\usepackage{amssymb}

\usepackage{url}
\usepackage{wasysym}
\usepackage{amsthm}
\usepackage{makecell}
\usepackage{diagbox}
\usepackage{subcaption}
\usepackage{algorithm}
\usepackage{algpseudocode}
\usepackage{multirow}
\usepackage{booktabs}
\usepackage[nocompress]{cite}

\usepackage{footnote}
\makesavenoteenv{tabular}
\makesavenoteenv{table}
\def\etal{\emph{et al}.}
\DeclareMathOperator{\dataaug}{data\_aug}
\DeclareMathOperator{\inverseaug}{inverse\_aug}

\newcommand{\red}[1]{\textcolor{red}{#1}}

\usepackage[breaklinks=true,bookmarks=false]{hyperref}

\cvprfinalcopy %

\def\cvprPaperID{8609} %

\ifcvprfinal\fi
\begin{document}

\title{Efficient Adversarial Training with Transferable Adversarial Examples}

\author{Haizhong Zheng \quad Ziqi Zhang \quad Juncheng Gu \quad Honglak Lee \quad Atul Prakash\\
University of Michigan, Ann Arbor\\
{\tt\small \{hzzheng, ziqizh, jcgu, honglak, aprakash\}@umich.edu}
}

\maketitle

\begin{abstract}
Adversarial training is an effective defense method to protect classification models against adversarial attacks.
However, one limitation of this approach is that it can require orders of magnitude additional training time due to high cost of generating strong adversarial examples during training.
In this paper, we first show that there is high transferability between models from neighboring epochs in the same training process, i.e., adversarial examples from one epoch continue to be adversarial in subsequent epochs. Leveraging this property, we propose a novel method, Adversarial Training with Transferable Adversarial Examples (ATTA), that can enhance the robustness of trained models and greatly improve the training efficiency by accumulating adversarial perturbations through epochs.
Compared to state-of-the-art adversarial training methods, ATTA enhances adversarial accuracy by up to $7.2\%$ on {CIFAR10} and requires $12\sim14\times$ less training time on MNIST and CIFAR10 datasets with comparable model robustness. Our code is publicized at \url{https://github.com/hzzheng93/ATTA} .
\end{abstract}

\input{01_introduction}

\input{02_preliminary}

\input{03_transferability}

\input{04_atta}

\input{05_experiment}

\input{06_relatedwork}

{\small
\bibliographystyle{ieee_fullname}
\bibliography{egbib}
}

\input{appendix}

\end{document}

%% file: 01_introduction.tex
\section{Introduction}
State-of-the-art deep learning models for computer vision tasks have been found to be vulnerable to adversarial examples~\cite{szegedy2013intriguing}. Even a small perturbation on the image can fool a well-trained model. Recent works~\cite{eykholt2017robust, song2018physical, sharif2016accessorize} also show that adversarial examples can be physically realized, which can lead to serious safety issues.
Design of robust models, which correctly classifies adversarial examples, is an active research area~\cite{cohen2019certified,lecun1998gradient,raghunathan2018certified,hein2017formal,wong2017provable}, with \emph{adversarial training}~\cite{madry2017towards} being one of the most effective methods.
It formulates training as a game between adversarial attacks and the model: the stronger the adversarial examples generated to train the model are, the more robust the model is.

To generate strong adversarial examples, iterative attacks~\cite{kurakin2016adversarial}, which use multiple attack iterations to generate adversarial examples, are widely adopted in various adversarial training methods\cite{madry2017towards, zhang2019theoretically, carmon2019unlabeled, tramer2017ensemble}. 
Since adversarial perturbations are usually bounded by a constrained space $\mathcal{S}$ and attack perturbations outside $\mathcal{S}$ need to be projected back to $\mathcal{S}$, $k$-step projected gradient descent method\cite{kurakin2016adversarial, madry2017towards} (PGD-$k$) has been widely adopted to generate adversarial examples. Typically, using more attack iterations (higher value of $k$) produces stronger adversarial examples~\cite{madry2017towards}.
However, each attack iteration needs to compute the gradient on the input, which causes a large computational overhead. As shown in Table~\ref{tab:adv-t-times}, the training time of adversarial training can be close to $100$ times larger than natural training.

Recent works~\cite{papernot2017practical, liu2016delving, papernot2016transferability} show that adversarial examples can be transferred between models: adversarial examples generated for one model can still stay adversarial to another model. 
The key insight in our work, which we experimentally verified, is that, because of high transferability between models (i.e., checkpoints) from neighboring training epochs, attack strength can be accumulated across epochs by repeatedly reusing the adversarial perturbation from the previous epoch.

We take advantage of this insight in coming up with a novel adversarial training method called ATTA (Adversarial Training with Transferable Adversarial examples) that can be significantly faster than state-of-the-art methods while achieving similar model robustness. 
In traditional adversarial training, when a new epoch begins, the attack algorithm generates adversarial examples from the original input images, which ignores the fact that these perturbations can be reused effectively. In contrast, we show that it can be advantageous to reuse these adversarial perturbations through epochs. 
Even using only {\em one attack iteration} to generate adversarial examples, ATTA-$1$ can still achieve comparable robustness with respect to traditional PGD-$40$.

We apply our technique on Madry's Adversarial Training method (MAT) \cite{madry2017towards} and TRADES\cite{zhang2019theoretically} and evaluate the performance on both MNIST and CIFAR10 dataset. 
Compared with traditional PGD attack, our method improves training efficiency by up to $12.2\times$ ($14.1\times$) on MNIST (CIFAR10) with comparable model robustness. Trained with ATTA, the adversarial accuracy of MAT can be improved by up to $7.2\%$. Noticeably, for MNIST, with one attack iteration, our method can achieve $96.31\%$ adversarial accuracy within $5$ minutes (compared to $65$ minutes with MAT ). For CIFAR10, compared to MAT whose training time is more than one day, our method achieves comparable adversarial accuracy in about $2$ hours (compared to $33$ hours with MAT). 
We also verify that our method is practical for large dataset: we evaluate our method with ImageNet, which can achieve comparative robustness within $97$ hours with just one 2080 TI GPU.
We also compare our method with two fast adversarial training methods, YOPO\cite{zhang2019you} and Free\cite{shafahi2019adversarial}. Evaluation results show that our method is both more effective and efficient than these two methods.

\textbf{Contribution.} To the best of our knowledge, our work is the first to enhance the efficiency and effectiveness of adversarial training by taking advantage of high transferability between models from different epochs.
In summary, we make the following contributions:
\begin{itemize}
  \item We are the first to reveal the high transferability between models of neighboring epochs in adversarial training. 
  With this property, we verify that the attack strength can be accumulated across epochs by reusing adversarial perturbations from the previous epoch.

  \item We propose a novel method (ATTA) for iterative attack based adversarial training with the objectives of both efficiency and effectiveness.
  It can generate the similar  (or even stronger) adversarial examples with much fewer attack iterations via accumulating adversarial perturbations through epochs.
  
  \item Evaluation result shows that, with comparable model robustness, ATTA is $12.2\times$ ($14.1\times$) faster than traditional adversarial methods on MNIST (CIFAR10). ATTA can also enhance model adversarial accuracy by up to $7.2\%$ for MAT on CIFAR10.
  
\end{itemize}

%% file: 02_preliminary.tex
\section{Adversarial training and transferability}
In this section, we introduce relevant background on adversarial training and transferability of adversarial examples. We also discuss the trade-off between training time and model robustness.

\subsection{Adversarial Training} \label{ssec:bg-adv-t}
Adversarial training is an effective defense method to train robust models against adversarial attacks.
By using adversarial attacks as a data augmentation method, a model trained with adversarial examples achieves considerable robustness.
Recently, lots of works\cite{madry2017towards, zhang2019theoretically, zhang2019limitations, cai2018curriculum, jang2019adversarial, hendrycks2019using, schmidt2018adversarially} focus on analyzing and improving adversarial machine learning. Madry \etal ~\cite{madry2017towards} first formulate adversarial training as a min-max optimization problem:
\begin{equation} \label{eq:adv-t}
 \min_{f \in \mathcal{H}} \mathbb{E}_{(x,y)\sim \mathcal{D}}[\max_{\delta \in \mathcal{S}} L(f(x+\delta), y)]
\end{equation}
where $\mathcal{H}$ is the hypothesis space, $\mathcal{D}$ is the distribution of the training dataset, $L$ is a loss function, and $\mathcal{S}$ is the allowed perturbation space that is usually selected as an L-$p$ norm ball around $x$.  The basic strategy in adversarial training is, given a original image $x$, to find a perturbed image $x+\delta$ that maximizes the loss with respect to correct classification. 
The model is then trained on generated adversarial examples. In this work, We consider adversarial examples with a high loss to have {\em high attack strength}.

\textbf{PGD-$k$ attack based adversarial training}: 
Unfortunately, solving the inner maximization problem is hard. 
Iterative attack \cite{kurakin2016adversarial} is commonly used to generate strong adversarial examples as an approximate solution for the inner maximization problem of Equation~\ref{eq:adv-t}. 
Since adversarial perturbations are usually bounded by the allowed perturbation space $\mathcal{S}$, PGD-$k$ ($k$-step projected gradient descent \cite{kurakin2016adversarial}) is adopted to conduct iterative attack \cite{madry2017towards, zhang2019theoretically, zhang2019you, shafahi2019adversarial}.
PGD-$k$ adversarial attack is the multi-step projected gradient descent on a negative loss function:
$$x_{t+1} = \Pi_{x+\mathcal{S}}(x_t + \alpha \text{ sgn}(\Delta_{x_t} L(\theta, x_t, y)))$$

In the above, $x_t$ is the adversarial example in the $t$-th attack iteration, $\alpha$ is the attack step size, and $\Pi$ is the projection function to project adversarial examples back to the allowed perturbation space $\mathcal{S}$.

With a higher value of $k$ (more attack iterations), PGD-$k$ can generate adversarial examples with higher loss \cite{madry2017towards}. 
However, there is a trade-off between training time and model robustness in adversarial training. 
On one hand, since each attack iteration needs to calculate the gradient for the input, using more attack iterations requires more time to generate adversarial examples, thus causing a large computational overhead for adversarial training. 
As shown in Table~\ref{tab:adv-t-times}, compared to natural training, adversarial training may need close to $100$x  more training time until the model converges.  Most of the training time is consumed in generating adversarial examples (attack time).
On the other hand, reducing the number of attack iterations $k$ can reduce the training time, but that negatively impacts robustness of the trained model (Table~\ref{tab:robustness-iteration}).

\begin{table}[h]
 \centering
 \begin{tabular}{ccccc}
 \toprule
 \multirow{2}{*}{Dataset} & \multirow{2}{*}{\begin{tabular}[c]{@{}c@{}}Natural \\ Training\end{tabular}} & \multicolumn{3}{c}{Adversarial training} \\ \cline{3-5}
 & & Training & Attack & Total \\ \midrule
 MNIST & $39.7$ sec & $210$ sec& $3723$ sec & $3933$ sec\\
 CIAFR10 & $55$min & $214$ min & $1813$ min & $2027$ min\\
 \bottomrule
 \end{tabular}
 \caption{Training time of natural training and adversarial training. {\em Attack} column shows the time consumed in adversarial example generation.}
 \label{tab:adv-t-times}
\end{table}

\begin{table}[h]
 \centering
 \begin{tabular}{ccccc}
 \toprule
 Defense & PGD-$1$ & PGD-$3$ & PGD-$5$ & PGD-$10$ \\ \midrule
 Nat. Acc. &$93.18\%$&$ 89.95\%$&$85.24\%$&$87.49\%$ \\
 Adv. Acc. &$22.3\%$&$41.38\%$&$43.55\%$&$47.07\%$ \\
 Time (min) &$435$&$785$&$1140$&$2027$ \\
 \bottomrule
 \end{tabular}
 \caption{The relation between the number of attack iterations and adversarial accuracy against PGD-$20$ attack on CIFAR10 dataset.} \label{tab:robustness-iteration}
\end{table}

\subsection{Transferability of Adversarial Examples.} \label{ssec:transferability}
Szegedy \etal \cite{szegedy2013intriguing} show that adversarial examples generated for one model can stay adversarial for other models. This property is named as \emph{transferability}.
This property is usually leveraged to perform a black-box attack \cite{papernot2016transferability,papernot2017practical,li2018learning, liu2016delving}. 
To attack a targeted model $f_t$, the attacker generates transferable adversarial examples from the source model $f_s$. The higher the transferability between $f_s$ and $f_t$ is, the higher the success rate the attack has.

\textbf{Substitute model training} is a commonly used method to train a source model $f_s$. Rather than the benchmark label $y$, $f_s$ is trained with $f_t(x)$ which is the prediction result of the targeted model~\cite{papernot2017practical} to achieve a higher black-box attack success rate.
While our work does not use black-box attacks, we do rely on a similar intuition as behind substitute model training, namely, two models with similar behavior and decision boundaries are likely to have higher transferability between each other. We use this intuition to show high transferability between models from neighboring training epochs, as discussed in the next section.

%% file: 03_transferability.tex
\section{Attack strength accumulation} \label{sec:trans}
In this section, we first conduct a study and find that models from neighboring epochs show very high transferability and are naturally good substitute models to each other. Based on this observation, we design an accumulative PGD-$k$ attack that accumulates attack strength by reusing adversarial perturbations from one epoch to the next. 
Compared to traditional PGD-$k$ attack, accumulative PGD-$k$ attack achieves much higher attack strength with fewer number of attack iterations in each epoch.

\subsection{High transferability between epochs}
Transferability between models in different training epochs of the same training program has not been studied.
Because fluctuations of model parameters between different epochs are very small,  we think that
they are likely to have similar behavior and similar decision boundaries, which should lead to high transferability between these models.

To evaluate the transferability between models from training epochs, we adversarially train a model $T$ as the targeted model, 
while saving intermediate models at the end of three immediately prior epochs as $T_1$, $T_2$, and $T_3$, with $T_3$ being the model from the epoch immediately prior to $T$. 
We measure the transferability of adversarial examples of each of $T_i$ with $T$. For comparison, we also train three additional models $S_1$, $S_2$, and $S_3$ that are trained by exactly the same training method as $T$ but with different random seeds. And we measure the transferability between each of $S_i$ and $T$.

To measure transferability from the source model to the targeted model, we use two metrics.  
The first metric is {\em error rate transferability} used in \cite{baluja2018learning, papernot2016transferability}, which is the ratio of the number of adversarial examples misclassified by source model to that of the targeted model.
The other metric is {\em loss transferability}, which is the ratio of the loss value caused by adversarial examples on the source model to the loss value caused by the same examples on the targeted model.

\begin{figure}[tb]
  \begin{subfigure}[b]{0.5\textwidth}
    \centering
      \includegraphics[height=2.6cm]{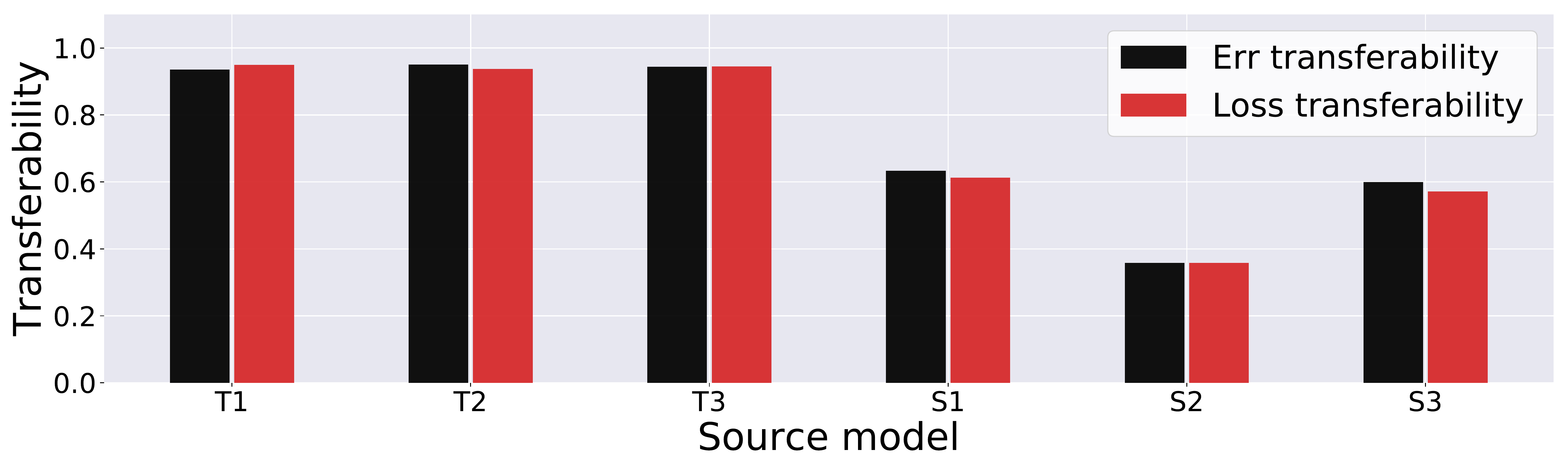}
      \caption{MNIST}
  \end{subfigure}
  \begin{subfigure}[b]{0.5\textwidth}
    \centering
      \includegraphics[height=2.6cm]{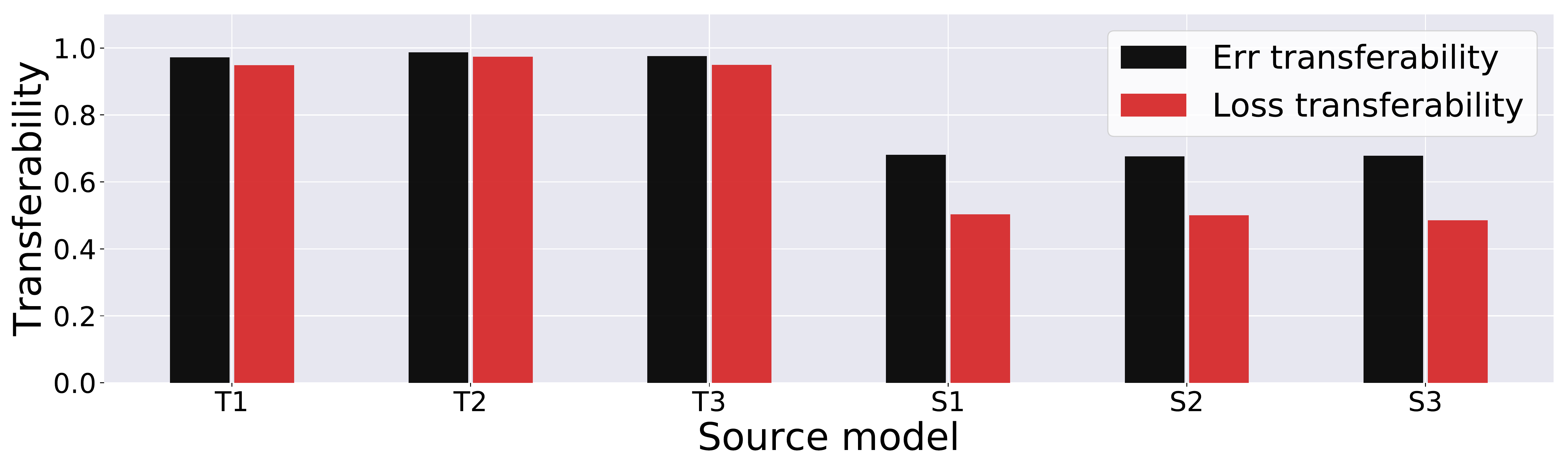}
      \caption{CIFAR10}
  \end{subfigure}
  \caption{Error rate transferability and loss transferability with the different source models.}
   \vspace{-0.3cm}
  \label{fig:transferability}
\end{figure}

We conduct experiments on both MNIST and CIFAR10 dataset, and the results are shown on Figure~\ref{fig:transferability}.
We find that, compared to the baseline models $S_i$, the models $T_i$ from neighboring epochs of $T$ have higher transferability for both transferability metrics (the transferability metrics for all models $T_i$ are larger than $0.93$). This provides strong empirical evidence that adversarial examples generated in one epoch still retain some strength in subsequent epochs.

Inspired by the above result, we state the following hypothesis.  {\bf Hypothesis:} {\em Repeatedly reusing perturbations from the previous epoch can accumulate attack strength epoch by epoch. Compared to current methods that iterate from original images in each epoch, this can allow us to use few attack iterations to generate the same strong adversarial examples.
}

\subsection{Accumulative PGD-$k$ attack} \label{ssec:acc-attack-eva}
\begin{figure}[ht]
 \centering
 \begin{subfigure}[b]{0.5\textwidth}
 \centering
 \includegraphics[height=1.7cm]{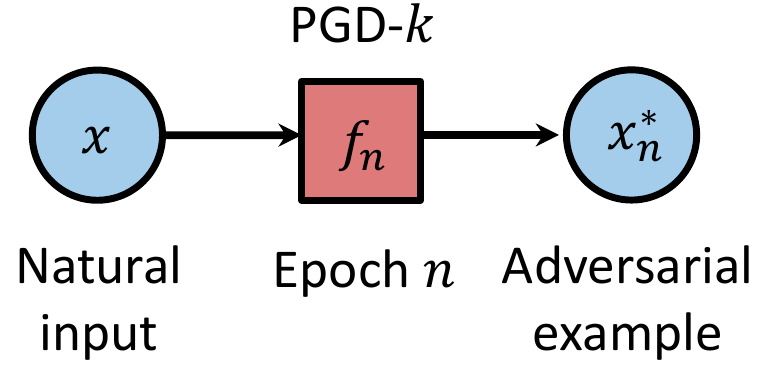}
 \caption{ } \label{fig:2a}
 \end{subfigure}
 \begin{subfigure}[b]{0.5\textwidth}
 \centering
 \includegraphics[height=2.1cm]{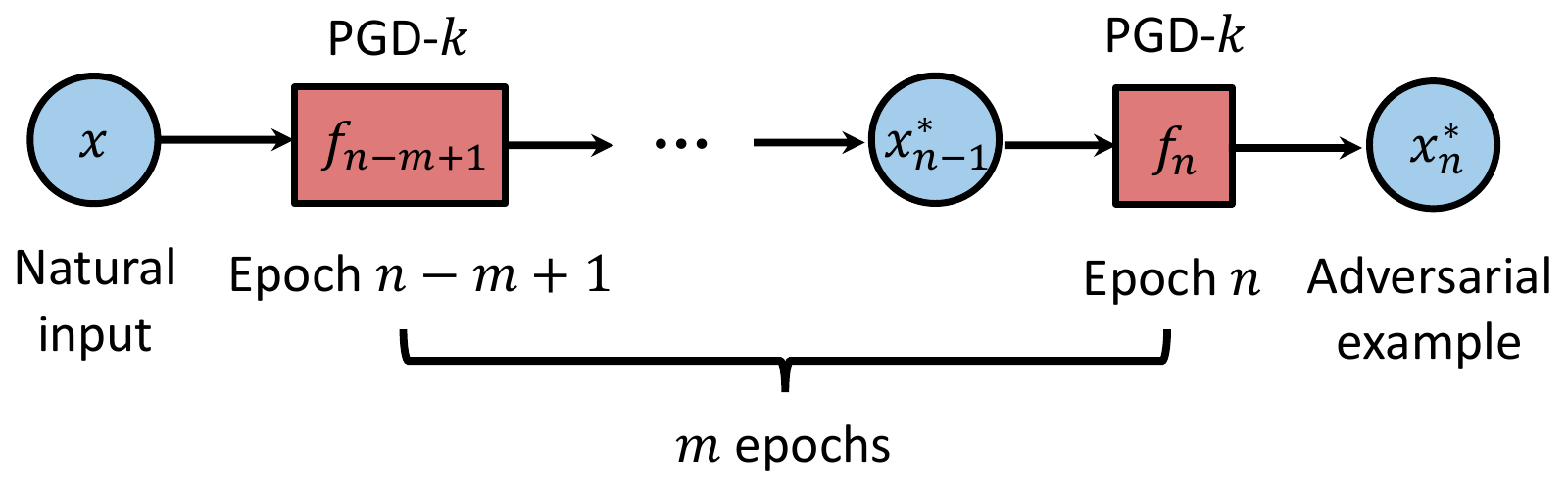}
 \caption{ } \label{fig:2b}
 \end{subfigure} 
 \caption{The traditional PGD-$k$ attack (a) and the accumulative PGD-$k$ attack through $m$ epochs (b). The PGD-$k$ attack is performed on the model in the red rectangle to get adversarial examples.}
 \label{fig:atta-attack}
\end{figure}

To validate the aforementioned hypothesis, we design an \emph{accumulative PGD-$k$ attack}.
As shown in Figure~\ref{fig:2b}, we longitudinally connect models in each epoch by directly reusing the attack perturbation of the previous epoch.
Accumulative PGD-$k$ attack Figure~\ref{fig:2b} generates adversarial examples for $f_{n-m+1}$ first. 
Then, for the following epochs, the attack is performed based on the accumulated perturbations from previous epochs.

\begin{figure}[ht]
 \begin{subfigure}[b]{0.5\textwidth}
 \centering
 \includegraphics[height=5cm]{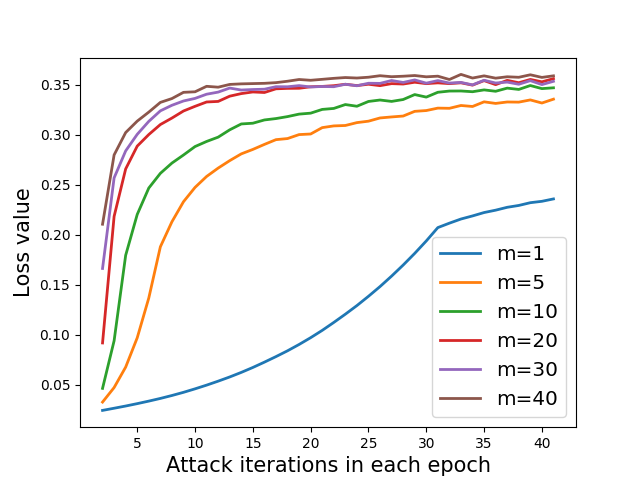}
 \caption{MNIST}
 \end{subfigure}
 \begin{subfigure}[b]{0.5\textwidth}
 \centering
 \includegraphics[height=5cm]{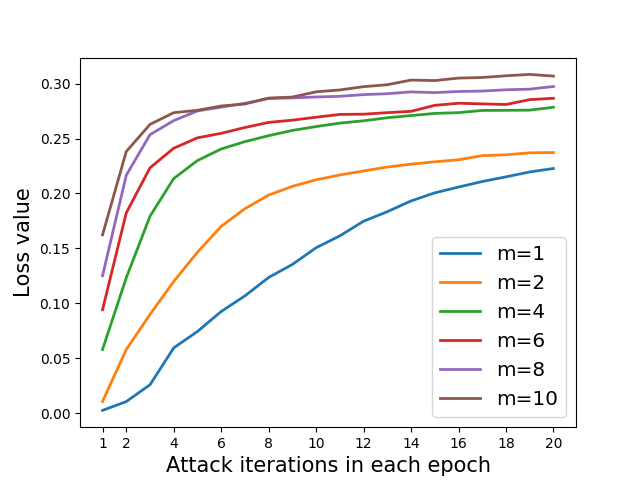}
 \caption{CIFAR10}
 \end{subfigure}
 \caption{
 Given the number of epochs $m$, the relationship between the number of attack iterations $k$ and the attack loss. $m=1$ stands for the traditional PGD-$k$ attack.}
 \label{fig:acc-attack-com}
\end{figure}

To compare the attack strength of two attacks, we use Madry's method \cite{madry2017towards} to adversarially train two models on MNIST and CIFAR10 and evaluate the loss value $\mathcal{L}(f_n(x^*_n), y)$ of adversarial examples generated by two attacks.
Figure~\ref{fig:acc-attack-com} summarises the evaluation result.
We can find that, with more epochs $m$ involved in the attack, accumulative PGD-$k$ attack can achieve a higher loss value with the same number of attack iterations $k$.

Especially, when adversarial examples are transferred through a large number of epochs, even accumulative PGD-$1$ attack can cause high attack loss. For MNIST, accumulative PGD-$1$ attack can achieve the same attack loss as traditional PGD-$35$ attack when $m=40$. For CIFAR10, accumulative PGD-$1$ attack can achieve the same attack loss as traditional PGD-$12$ attack when $m=10$.

This result indicates that, with high transferability between epochs, adversarial perturbations can be reused effectively, which allows us to use fewer attack iterations to generate the same or stronger adversarial examples. Reuse of perturbations across epochs can help us reduce the number of attack iterations $k$ in PGD-$k$, leading to more efficient adversarial training. Next section describes our proposed algorithm, \textbf{ATTA} (Adversarial Training with Transferable Adversarial examples), based on this property.

%% file: 04_atta.tex
\section{Adversarial training with transferable adversarial examples}

\begin{figure}[htbp!]
 \centering
 \includegraphics[width=0.47\textwidth]{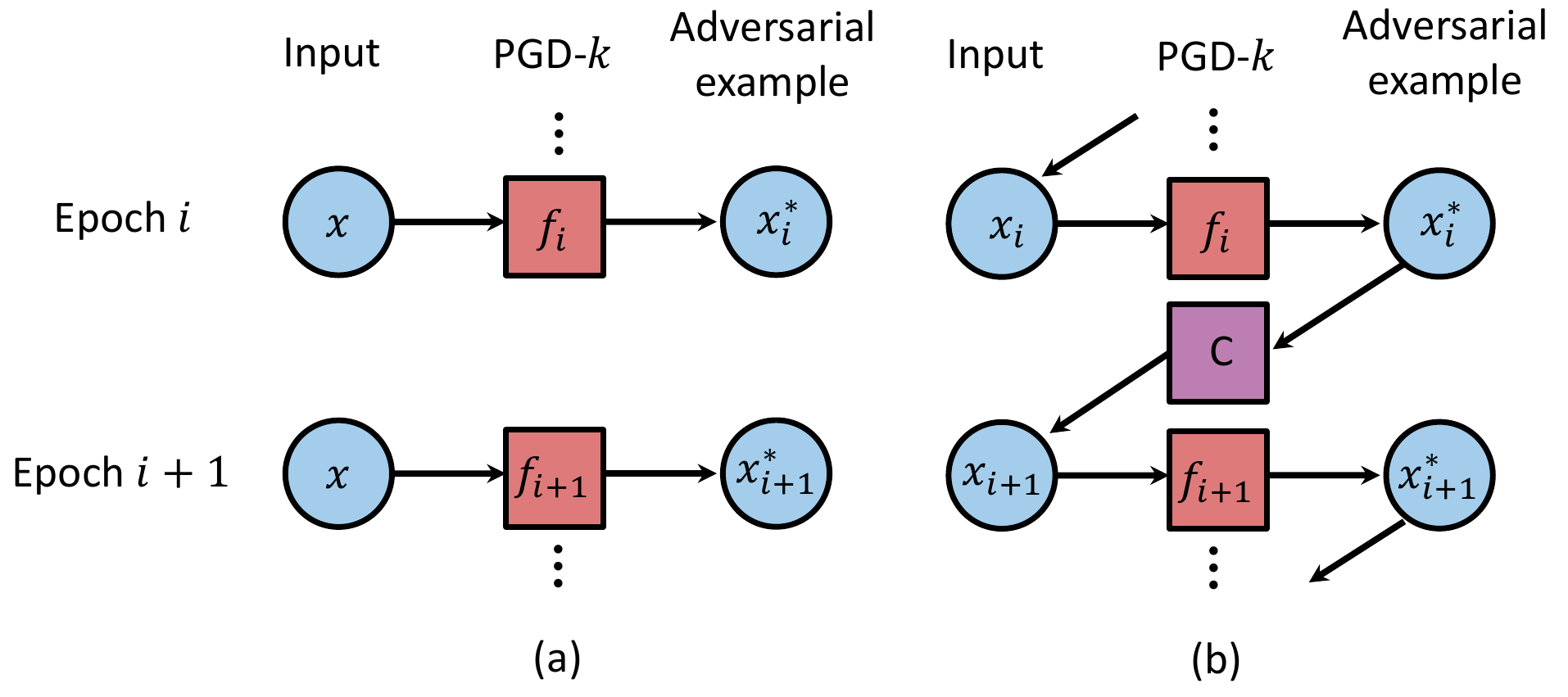}
 \caption{Traditional adversarial training (PGD-$k$) (a) and ATTA-$k$ (b). $C$ is the connection function which improves the transferability between epochs.}
 \label{fig:pgd-atta-comparison}
\end{figure}

The discussion on transferability in Section~\ref{sec:trans} suggests that adversarial examples can retain attack strength in subsequent training epochs.
The results of accumulative attack in Section~\ref{sec:trans} suggest that stronger adversarial examples can be generated by accumulating attack strength.
This property inspires us to \emph{link} adversarial examples between adjacent epochs as shown in Figure~\ref{fig:pgd-atta-comparison}. Unlike starting from a natural image $x$ to generate an adversarial example in each epoch as shown in Figure~\ref{fig:pgd-atta-comparison}(a), we start from a previously saved adversarial example ${x_i}^*$ from the previous epoch to generate an adversarial example (Figure~\ref{fig:pgd-atta-comparison}(b)). To improve the transferability between epochs, we use a connection function $C$ to link adversarial examples, which transforms $x^*_i$ to a start point $x_{i+1}$ for the next epoch.
During the training, with repeatedly reusing adversarial example between epochs, attack strength can be accumulated epoch by epoch:
$$x^*_{i+1} = \mathcal{A}(f_{i+1}, x, y, C(x^*_{i}))$$
where $\mathcal{A}$ is the attack algorithm, $C$ is a connection function (described in the next section). $f_i$ is the model in the $i$-th epoch, $x^*_{i}$ is the adversarial examples generated in the $i$-th epoch and $x,y$ are natural image and benchmark label \footnote{Note that the $x^*_{i+1}$ is still bounded by the natural image $x$, rather than $C(x^*_{i})$.}.

As shown in the previous section, adversarial examples can achieve high attack strength as they are transferred across epochs via the above linking process, rather than starting from natural images. This, in turn, should allow us to train a more robust model with fewer attack iterations.

\subsection{Connection function design} \label{ssec:connection-function}
Designing connection function can help us overcome two challenges that we encountered in achieving high transferability of adversarial examples from one epoch to the next: 
\begin{enumerate}

\item {\em Data augmentation problem:} Data augmentation is a commonly used technique to improve the performance of deep learning.
It applies randomly transformation on original images so that models can be trained with various images in different epochs. This difference can cause a mismatch of images between epochs.
Since the perturbation is calculated with the gradient of image, if we directly reuse the perturbation, the mismatch between reused perturbation and new augmented image can cause a decrease in attack strength. Simply removing data augmentation also hurts the robustness. We experimentally show this negative influence of these two alternatives in Section~\ref{ssec:ablation-study}.

\item {\em Drastic model parameter change problem:} As discussed in Section~\ref{sec:trans}, similar parameters between models tends to cause a similar decision boundary and thus high transferability. Unfortunately, model parameters tend to change drastically at the early stages of training.
Thus, adversarial perturbations in early epochs tend to be useless for subsequent epochs.
\end{enumerate}

\begin{figure}[!t]
 \centering
 \includegraphics[width=0.48\textwidth]{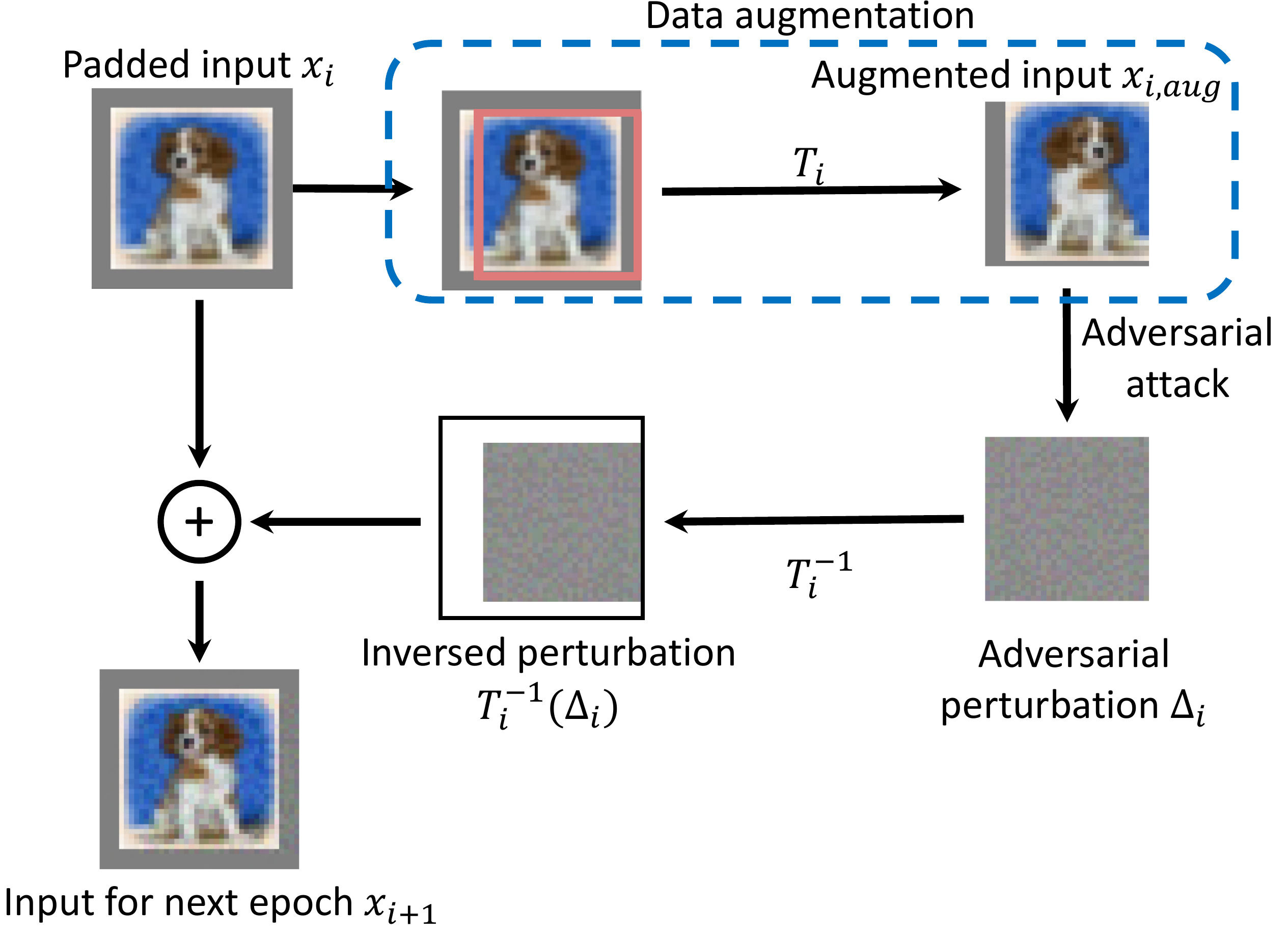}
 \caption{The workflow of inversed data augmentation.}
 \label{fig:aug}
\end{figure}

\textbf{Overcoming challenges: Inverse data augmentation.}
To address the first issue, we propose a technique called {\em inverse data augmentation}
so that adversarial examples retain a high transferability between training epochs despite data augmentation. Figure~\ref{fig:aug} shows the workflow with inverse data augmentation. Some transformations (like cropping and rotation) pad augmented images with background pixels. To transfer a perturbation on background pixels, our method transfers the padded image $x_i$ rather than standard images so that the background pixels used by data augmentation can be covered by these paddings.

After the adversarial perturbation of augmented image is generated,
we can perform the inverse transformation\footnote{In adversarial training, most data augmentation methods used are linear transformations which are easy to be inversed.} $T^{-1}(\cdot)$ to calculate the inversed perturbation $T^{-1}(\Delta_i)$ on the padded image $x_i$. By adding $T^{-1}(\Delta_i)$ to $x_i$, we can store and transfer all perturbation information in $x_{i+1}$ to next epoch.
(Note that, when we perform the adversarial attack on the augmented image $x_{i,aug}$, the perturbation is still bounded by the natural image rather than $x_{i,aug}$.)

\textbf{Periodically reset perturbation.}
To solve the second issue, we propose a straightforward but effective solution: our method resets the perturbation and lets adversarial perturbations be accumulated from the beginning periodically, which mitigates the impact caused by early perturbations.

\subsection{Attack Loss}
Various attack loss functions are used in different adversarial training algorithms. For example, TRADES\cite{zhang2019theoretically} uses robustness loss (Equation~\ref{eq:trades-attack-loss}) as the loss to generate the adversarial examples.
\begin{equation}\label{eq:trades-attack-loss}
 \mathcal{L}(f(x^*), f(x))
\end{equation}
where $\mathcal{L}$ is the loss function, $f$ is the model and $x$, $x^*$ are natural and adversarial example respectively.
This loss represents how much the adversarial example diverges from the natural image. Zhang \etal \cite{zhang2019theoretically} shows that this loss has a better performance in the TRADES algorithm.

In our method, we use the following loss function:
\begin{equation}\label{eq:madry-attack-loss}
 \mathcal{L}(f(x^*), y)
\end{equation}
It represents how much the adversarial examples diverges to the benchmark label.

The objection ($f(x)$) of Equation~\ref{eq:trades-attack-loss} for adversarial attack varies across epochs, which may weaken the transferability between epochs. Equation~\ref{eq:madry-attack-loss} applies a fixed objection ($y$), which doesn't have this concern.
In addition, Equation~\ref{eq:madry-attack-loss} has a smaller computational graph, which can reduce computational overhead during training.

The overall training method is described in Algorithm~\ref{alg:atta}.

\begin{algorithm}[th]
 \caption{Adversarial Training with Transferable Adversarial Examples (ATTA)}\label{alg:atta}
 \begin{algorithmic}[1]
 \State \textbf{Input}: Padded training dataset $D_{nat}$, model $f_\theta$, attack algorithm $\mathcal{A}$, perturbation bound $\epsilon$, the number of epochs to reset perturbation $reset$
 \State Initialize $\theta$
 \State Initialize $D$ by cloning $D_{nat}$
 \For{$epoch = 1 \cdots N$}
 \For{$x_{nat},y$ in $D_{nat}$ and corresponding $x \in D$}
 \If {$epoch\ \%\ reset\ =\ 0$}
 \State $\beta \gets$ a small random perturbation
 \State $x \gets x_{nat} + \beta$
 \EndIf
 \State Store the transformation $T_{aug}$ for the inverse augmentation:
 \State $x_{aug}, x_{nat, aug}, T_{aug} \gets \dataaug(x, x_{adv})$
 \State $x^* \gets \mathcal{A}(f_{\theta}, x_{nat, aug}, y, x_{aug}, \epsilon)$
 \State $\theta \gets \theta - \nabla_{f_{\theta}} \frac{\partial \mathcal{L}(f_\theta, x^*, y)}{\partial \theta}$
 \State $x \gets \inverseaug(x, x^*, T_{aug})$
 \EndFor
 \EndFor
 \end{algorithmic}

\end{algorithm}

%% file: 05_experiment.tex
\section{Evaluation}
In this section, we integrate ATTA with two popular adversarial training methods: Madry's Adversarial Training (MAT) \cite{madry2017towards} and TRADES \cite{zhang2019theoretically}. By evaluating the training time and robustness, we show that ATTA can provide a better trade-off than other adversarial training methods.
To understand the contribution of each component to robustness, we conduct an ablation study.

\subsection{Setup}
Following the literature~\cite{madry2017towards, zhang2019theoretically, zhang2019you}, we use both MNIST~\cite{lecun1998gradient} and CIFAR10 dataset~\cite{krizhevsky2009learning} to evaluate ATTA.

For the MNIST dataset, the model has four convolutional layers followed by three full-connected layers which is same architecture as used in \cite{madry2017towards, zhang2019theoretically}. The adversarial perturbation is bounded by $l_\infty$ ball with size $\epsilon = 0.3$.

For the CIFAR10 dataset, we use the wide residual network WRN-34-10 \cite{zagoruyko2016wide} which is same as \cite{madry2017towards, zhang2019theoretically}. The perturbation is bounded by $l_\infty$ ball with size $\epsilon = 0.031$.

\subsection{Efficiency and effectiveness of ATTA}
In this part, we evaluate the training efficiency and the robustness of ATTA, comparing it to state-of-the-art adversarial training algorithms. To better verify effectiveness of ATTA, we also evaluate ATTA under various attacks.

\subsubsection{Training efficiency}
We select four state-of-the-art adversarial training methods as baselines: MAT\cite{madry2017towards}, TRADES\cite{zhang2019theoretically}, YOPO\cite{zhang2019you} and Free\cite{shafahi2019adversarial}.
For MNIST, the model trained with ATTA can achieve a comparable robustness with up to $12.2$ times training efficiency and,
for CIFAR10, our method can achieve a comparable robustness with up to $14.1$ times training efficiency. Compared to MAT trained with PGD, our method improve the accuracy up to $7.2\%$ with $3.7$ times training efficiency.

\renewcommand{\arraystretch}{1.3}

\begin{table}[h]
  \centering
  \begin{tabular}{c*{3}{c}c}\toprule
  \multicolumn{2}{c}{\backslashbox{Defense}{Attack}}&Natural&PGD-40&\multicolumn{1}{c}{\begin{tabular}[c]{@{}c@{}}Time\\ (sec)\end{tabular}}\\\midrule
  \multirow{5}{*}{MAT}
  &PGD-1 & $99.52\%$&$15.82\%$&$226$\\
  &PGD-40 & $\textbf{99.37\%}$&$\textbf{96.21\%}$&$\textbf{3933}$\\\cmidrule{2-5}
  &YOPO-$5$-$10$ &$99.15\%$&$93.69\%$&$789$\\\cmidrule{2-5}
  &ATTA-1 &$\textbf{99.45\%}$&$\textbf{96.31\%}$&$\textbf{297}$\\
  &ATTA-40 &$99.23\%$&$97.28\%$&$4650$\\
  \midrule\midrule

  \multirow{4}{*}{TRADES}
  &PGD-1 &$99.41\%$&$39.53\%$&$583$\\
  &PGD-40 &$\textbf{98.89\%}$&$\textbf{96.54\%}$&$\textbf{6544}$\\\cmidrule{2-5}
  &ATTA-1 &$\textbf{99.03\%}$&$\textbf{96.10\%}$&$\textbf{460}$\\
  &ATTA-40 &$98.21\%$&$96.03\%$&$4660$\\
  \bottomrule
  \end{tabular}
  \caption{The result of different attacks on MNIST dataset.}
  \label{tab:exp-mnist}
\end{table}

\textbf{MNIST.} The experiment results of MNIST are summarised in Table~\ref{tab:exp-mnist}. For MAT, to achieve comparable robustness, ATTA is about $12.2$ times faster than the traditional PGD training method. Even with one attack iteration, model trained with ATTA-$1$ achieves $96.31\%$ adversarial accuracy within $297$ seconds.
For TRADES, we get a similar result. With one attack iteration, ATTA-$1$ achieves $96.1\%$ adversarial accuracy within $460$ seconds, which is about $13$ times faster than TRADES(PGD-$40$). Compared to another fast adversarial training method YOPO, ATTA-$1$ is about $2.7$ times faster and achieves higher robustness ($96.1\%$ versus YOPO's $93.69\%$).

\begin{table}[h]
  \centering
  \begin{tabular}{c*{3}{c}c}\toprule
  \multicolumn{2}{c}{\backslashbox{Defense}{Attack}}&Natural&PGD-20&\multicolumn{1}{c}{\begin{tabular}[c]{@{}c@{}}Time\\ (min)\end{tabular}}\\\midrule
  \multirow{8}{*}{MAT}
  &PGD-1 &$93.18\%$&$22.3\%$&$435$\\
  &PGD-3 &$89.95\%$&$41.38\%$&$785$\\
  &PGD-10 & $\textbf{87.49\%}$& $\textbf{47.07\%}$&$\textbf{2027}$\\\cmidrule{2-5}
  &Free($m=8$) &$85.54\%$&$47.68\%$&$640$\\
  &YOPO-$5$-$3$ &$86.43\%$&$48.24\%$&$335$\\\cmidrule{2-5}
  &ATTA-1 &$\textbf{85.71\%}$&$\textbf{50.96\%}$&$\textbf{134}$\\
  &ATTA-3 &$85.44\%$&$52.56\%$&$267$\\
  &ATTA-10 & $83.80\%$&$54.33\%$&$690$\\
  \midrule\midrule

  \multirow{7}{*}{TRADES}
  &PGD-1 &$93.58\%$&$35.52\%$&$592$\\
  &PGD-3 &$88.52\%$&$54.20\%$&$861$\\
  &PGD-10 &$\textbf{84.13\%}$&$\textbf{56.6\%}$&$\textbf{2028}$\\\cmidrule{2-5}
  &YOPO-$3$-$4$ \footnotemark &$87.82\%$& $46.13\%$& -\\\cmidrule{2-5}
  &ATTA-1 &$85.04\%$&$54.50\%$&$199$\\
  &ATTA-3 &$\textbf{84.23\%}$&$\textbf{56.36\%}$&$\textbf{320}$\\
  &ATTA-10 &$83.67\%$&$57.34\%$&$752$\\
  \bottomrule
  \end{tabular}
  \caption{The result of different attacks on CIFAR10 dataset.}
  \label{tab:exp-cifar}
\end{table}

\definecolor{ATTA}{RGB}{194, 79, 85}
\definecolor{PGD}{RGB}{78, 116, 174}
\definecolor{Free}{RGB}{219, 131, 87}
\definecolor{YOPO}{RGB}{88, 167, 106}

\textbf{CIFAR10.}\ \ \ We summarise the experiment results of CIFAR10 in Table~\ref{tab:exp-cifar}.
For MAT, compared to PGD-$10$, ATTA-$1$ achieves $3.89\%$ higher adversarial accuracy with about $15.1$ times training efficiency, and ATTA-$10$ improves adversarial accuracy by $7.26\%$ with $2.9$ times training efficiency when the model is trained with $10$ attack iterations.
For TRADES, ATTA-$3$ achieves comparable adversarial accuracy to PGD-$10$ with $5.3$ times faster training efficiency.
By comparing the experiment results with YOPO and Free, for MAT, our method is $3.78$ ($1.5$) times faster than Free (YOPO) with $3.28\%$ ($2.72\%$) better adversarial accuracy.

\begin{figure}[!t]
  \centering
  \includegraphics[width=0.47\textwidth]{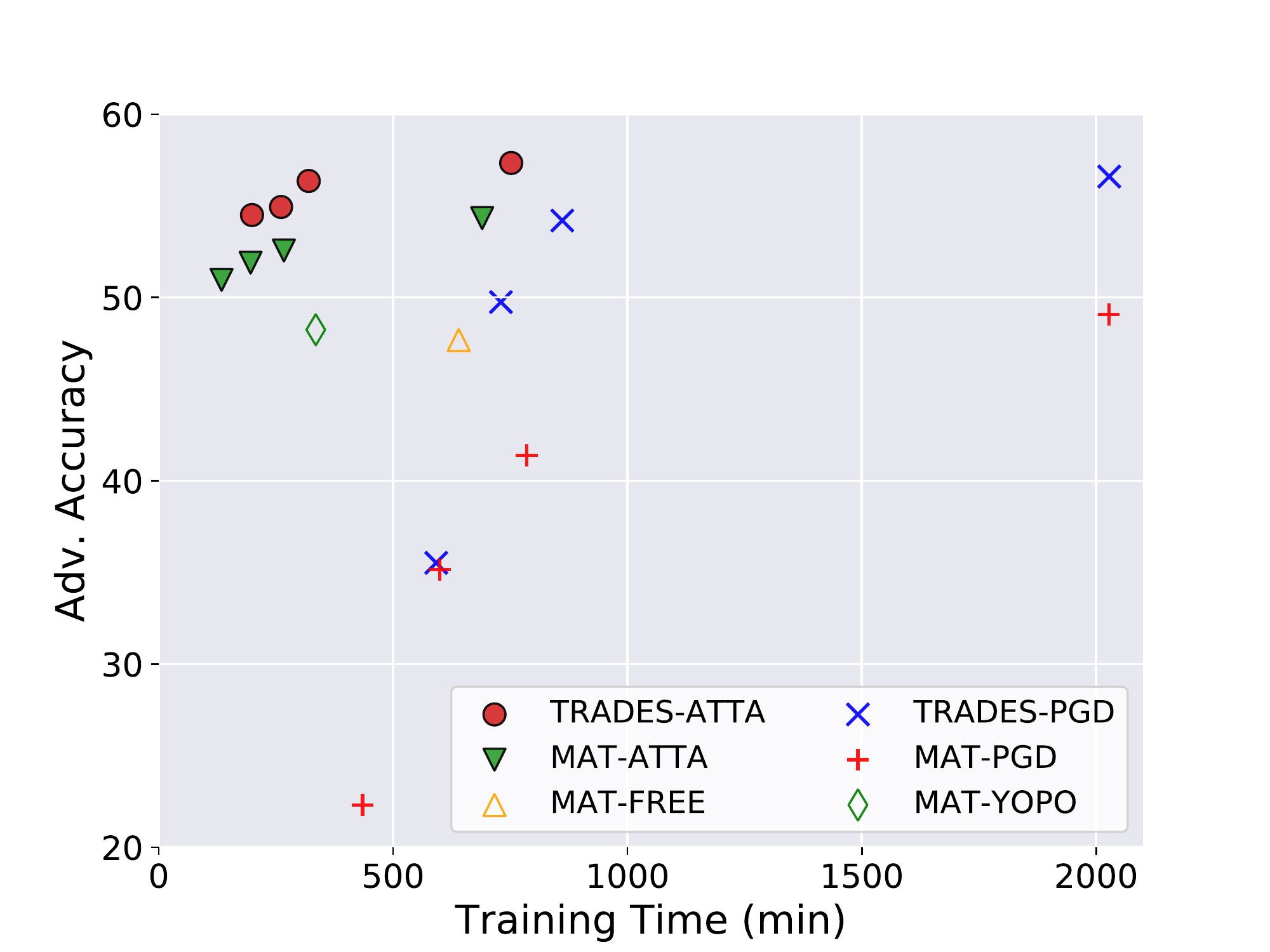}
  \caption{The scatter plot presents adversarial accuracy against PGD-$20$ attack and training time of different adversarial training methods on CIFAR10.
  }
  \label{fig:cifar-time-robustness}
\end{figure}

To better understand the performance of ATTA, we present the trade-off between training time and robustness of different methods in Figure~\ref{fig:cifar-time-robustness}. We find that our method (indicated by the solid markers in the left-top corner) gives a better trade-off on efficiency and effectiveness on adversarial training.

\footnotetext{The author-implemented YOPO-$3$-$4$ can't converge in our experiment. We pick the accuracy data from YOPO paper.}

\subsubsection{Defense under other attacks}
To verify the robustness of our method, we evaluate models in the previous section with other attacks: PGD-$100$\cite{kurakin2016adversarial}, FGSM\cite{goodfellow2014explaining}, CW-$20$\cite{carlini2017adversarial}.

\begin{table}[h]
  \centering
  \begin{tabular}{ccccc}
  \toprule
  Defense& PGD$100$ & FGSM & CW$20$ \\ \midrule
  \multicolumn{5}{c}{MNIST} \\ \midrule
  M-PGD-$40$ &$94.69\%$& $97.37\%$& $99.06\%$\\
  M-ATTA-$40$ &$96.85\%$& $98.55\%$& $98.02\%$\\
  \midrule
  \multicolumn{5}{c}{CIFAR10} \\ \midrule
  M-PGD-$10$ &$46.77\%$& $63.5\%$&$56.7\%$ &\\
  M-ATTA-$10$ &$52.6\%$& $63.49\%$& $75.37\%$ \\
  T-PGD-$10$ &$55.36 \%$& $63.02\%$ &$79.4\%$\\
  T-ATTA-$10$ & $56.39\%$& $64.1\%$ &$82.01\%$\\
  \bottomrule
  \end{tabular}
  \caption{The robustness comparison between ATTA and PGD under other
  attacks. The first `M' and `T' stand for MAT and TRADES, respectively.}\label{tab:various-attacks}
\end{table}

As shown in Table~\ref{tab:various-attacks}, models trained with ATTA are still robust to other attacks. Compared to baselines, our method still achieves a comparable or better robustness under other attacks.
We find that, although ATTA-$40$ has a similar robustness to PGD-$40$ under PGD-$20$ attack, with a stronger attack (e.g. PGD-$100$), ATTA-$40$ shows a better robustness .%

\subsubsection{ATTA with ImageNet} \label{sssec:imagenet}
Table~\ref{tab:imagenet} shows the results for training an ImageNet Classifier~\cite{imagenet_cvpr09} with ATTA
for $l_\infty(\epsilon = 2/255)$ attack with the loss used in MAT. We also show
previously reported results of Free ($m=4$) from~\cite{shafahi2019adversarial}.
Shafahi \etal~\cite{shafahi2019adversarial} report requiring about $50$ hours with four P100s to adversarially train their model.  ATTA required
$97$ hours with one 2080Ti.

\begin{table}[h]
  \centering
  \begin{tabular}{ccccc}
  \toprule
  Defense& Natural & PGD$10$ & PGD$50$ & PGD$100$ \\ \midrule
  Free($m=4$) &$64.44\%$& $43.52\%$& $43.39\%$ & $43.40\%$\\
  ATTA-$2$ &$60.70\%$& $44.57\%$& $43.57\%$ & $43.51\%$\\
  \bottomrule
  \end{tabular}
  \caption{Evaluation of ATTA on ImageNet}\label{tab:imagenet}
\end{table}

\subsection{Ablation study} \label{ssec:ablation-study}
To study the contribution of each component to robustness, we do an ablation study on inverse data augmentation and different attack loss functions.

\textbf{Inverse data augmentation.}
To study the robustness gain of inverse data augmentation(i.d.a.), we use ATTA-$1$ to adversarially train models by reusing the adversarial perturbation directly. As shown in Table~\ref{tab:abl-inv-aug}, for both MAT (ATTA-$1$) and TRADES (ATTA-$1$), models trained with inverse data augmentation achieve about $6\%$ higher accuracy, which means that inverse data augmentation does help improve the transferability between training epochs.
As discussed in \ref{ssec:connection-function}, another alternative is to remove data augmentation. However, Table~\ref{tab:abl-inv-aug} shows that removal of data augmentation hurts both natural accuracy and robustness.

\begin{table}[h]
 \centering
 \begin{tabular}{cccc}
 \toprule
 \backslashbox{Defense}{Attack}& Natural & PGD-$20$ &\\ \midrule
 MAT(w/o d.a., w/o i.d.a.) &$82.53\%$& $41.64\%$ \\
 MAT(w/ d.a., w/o i.d.a.) &$91.65\%$& $42.55\%$ \\
 MAT(w/ d.a., w/ i.d.a.) &$85.71\%$& $50.96\%$ \\
 TRADES(w/o d.a., w/o i.d.a.) &$81.87\%$&$41.65\%$  \\
 TRADES(w/ d.a., w/o i.d.a.) & $90.46\%$&$48.38\%$ \\
 TRADES(w/ d.a., w/ i.d.a.) &$85.04\%$&$54.50\%$  \\\bottomrule
 \end{tabular}
 \caption{The accuracy under PGD attack of models trained by ATTA-$1$ with or without d.a.(data augmentation) and i.d.a. (inverse data augmentation).}\label{tab:abl-inv-aug}
\end{table}

\textbf{Attack loss.} Zhang \etal \cite{zhang2019theoretically} show that, for TRADES, using Equation~\ref{eq:trades-attack-loss} leads to a better robustness. However, in this attack loss, we noticed that both inputs to the loss function $\mathcal{L}$ are related to the model $f$. Since the model $f$ is updated every epoch, compared to Equation~\ref{eq:madry-attack-loss} whose $y$ is fixed, the instability of $f(x^*)$ and $f(x)$ may have a larger influence on transferability. To analyze the performance difference between these two attack losses in ATTA, we train two TRADES(ATTA-$1$) models with different attack losses.
In Table~\ref{tab:abl-att-loss}, we find that Equation~\ref{eq:madry-attack-loss} leads to $2.48\%$ higher accuracy against PGD-$20$ attack. This result suggests that the higher stability of Equation~\ref{eq:madry-attack-loss} helps ATTA increase transferability between training epochs.

\begin{table}[h]
 \centering
 \begin{tabular}{cccc}
 \toprule
 \backslashbox{Defense}{Attack}& Natural & PGD-$20$\\ \midrule
 TRADES loss &$85.95\%$& $52.08\%$ \\
 MAT loss &$85.04\%$&$54.50\%$ \\ \bottomrule
 \end{tabular}
 \caption{The accuracy under PGD attack of models trained by TRADES(ATTA-$1$) with MAT loss and TRADES loss.}\label{tab:abl-att-loss}
\end{table}

%% file: 06_relatedwork.tex
\section{Related work}
Several works \cite{wang2019bilateral, cai2018curriculum, tramer2017ensemble, shafahi2018universal, carmon2019unlabeled, hendrycks2019using, madry2017towards, lee2017generative, zhang2019you, shafahi2019adversarial} focus on enhancing either the efficiency or effectiveness of the
adversarial training that was first proposed in~\cite{kurakin2016adversarial}. We highlight most
relevant work here. YOPO~\cite{zhang2019you}. TRADES~\cite{zhang2019theoretically} improves the robustness of an adversarially trained model by adding a robustness regularizer in the loss function. BAT~\cite{wang2019bilateral} uses perturbed labels (besides perturbed images) to generate strong attacks with fewer iterations than Madry \etal~\cite{madry2017towards}. ATTA does not perturb labels and is based on a completely different insight of transferability.

Free~\cite{shafahi2019adversarial} speeds up adversarial training by
recycling gradients calculated during training when generating adversarial examples.  In contrast, ATTA speedups adversarial training with a
completely different insight based on transferability.
While recycling gradients,
Free does reuse perturbations through iterations on the top of the same batch of images (inner-batch). In contrast, ATTA reuses perturbations {\em across} epochs (inter-epoch). Inter-epoch reuse is a non-trivial problem involving new challenges discussed in Section~4.1, requiring use of connection functions to overcome them. Free did not discuss transferability, but our analysis of transferability potentially provides an alternative way of understanding Free.
For many training scenarios, ATTA both outperforms existing techniques
as well as achieves better robustness.

\section{Conclusion}
ATTA is a new method for iterative attack based adversarial training that significantly speeds up training time and improves model robustness.
The key insight behind ATTA is high transferability between models from neighboring epochs, which is firstly revealed in this paper.
Based on this property, the attack strength in ATTA can be accumulated across epochs by repeatedly reusing adversarial perturbations from the previous epoch.
This allows ATTA to generate the same (or even stronger) adversarial examples with fewer attack iterations.
We evaluate ATTA and compare it with state-of-the-art adversarial training methods. It greatly shortens the training time with comparable or even better model robustness.
More importantly, ATTA is a generic method and can be applied to enhance the performance of other iterative attack based adversarial training methods.

\section{Acknowledgments}
We thank Fan Lai, Rui Liu, Jimmy You, Peifeng Yu, and Yiwen Zhang for helpful discussions. We thank
Ryan Feng and Renuka Kumar for feedback on paper drafts. This work is supported by NSF Grant No.1646392.

%% file: appendix.tex
\appendix

\section{Overview}
This supplementary material provides details on our experiment and additional evaluation results. In Section~\ref{sec:exp-setup}, we introduce the detailed setup of our experiment. In Section~\ref{sec:exp-details}, we compare adversarial examples generated by ATTA and PGD and show that, even with one attack iteration, ATTA-$1$ can generate similar perturbations to PGD-$40$ (PGD-$10$) on MNIST (CIFAR10). We also provide the complete evaluation results in Section~\ref{ssec:com-result}.

\section{Experiment setup} \label{sec:exp-setup}
We provide additional details on the implementation, model architecture, and hyper-parameters used in this work.

\textbf{MNIST.} We use the same model architecture used in \cite{madry2017towards, zhang2019theoretically, zhang2019you}, which has four convolutional layers followed by three fully-connected layers. The adversarial examples used to train the model are bounded by $l_\infty$ ball with size $\epsilon = 0.3$ and the step size for each attack iteration is $0.01$. We do not apply any data augmentation (and inverse data augmentation) on MNIST and set the epoch period to reset perturbation as infinity which means that perturbations are not reset during the training. The model is trained for $60$ epochs with an initial $0.1$ learning rate and a $0.01$ learning rate after $55$ epochs, which is the same as \cite{zhang2019you}.
To evaluate the model robustness, we perform the PGD~\cite{kurakin2016adversarial}, M-PGD~\cite{dong2018boosting} and CW~\cite{carlini2017adversarial} attack with a $0.01$ step size and set decay factor as $1$ for M-PGD (momentum PGD).

\textbf{CIFAR10.} Following other works~\cite{madry2017towards, zhang2019theoretically, zhang2019you, shafahi2019adversarial}, we use Wide-Resnet-34-10~\cite{zagoruyko2016wide} as the model architecture. The adversarial examples used to train the model are bounded by $l_\infty$ ball with size $\epsilon = 0.031$.
For ATTA-$1,2,3$, we use $0.015,0.01,0.01$ as the step size, respectively.
For ATTA-$k$ ($k>3$), we use $0.007$ as the step size.
The data augmentation used is a random flip and $4$-pixel padding crop, which is same with other works~\cite{madry2017towards, zhang2019theoretically, zhang2019you, shafahi2019adversarial}. We set the epoch period to reset perturbation as $10$ epochs. Following YOPO~\cite{zhang2019you}, the model is trained for $40$ epochs with an initial $0.1$ learning rate, a $0.01$ learning rate after $30$ epochs, and a $0.001$ learning rate after $35$ epochs.
To evaluate the model robustness, we perform the PGD, M-PGD and CW attack with a $0.003$ step size and set decay factor as $1$ for M-PGD (momentum PGD).

\textbf{ImageNet.} We use Resnet-50 \cite{he2016deep} as the model architecture. The adversarial examples used to train the model are bounded by $l_\infty$ ball with size $\epsilon = 2/255$, and we use we use $1/255$ as the step size. The model is trained for $20$ epochs with a piece-wise learning rate which is $0.4$ when epoch is less than $6$, $0.04$ when epoch is between $6$ and $11$, $0.004$ when epoch is between $12$ and $14$, $0.0004$ when epoch is larger than $14$.

For the baseline, we use the author implementation of MAT\footnote{
\url{https://github.com/MadryLab/mnist_challenge} \\ \url{https://github.com/MadryLab/cifar10_challenge}}~\cite{madry2017towards}, TRADES\footnote{\url{https://github.com/yaodongyu/TRADES}}~\cite{zhang2019theoretically}, YOPO\footnote{\url{https://github.com/a1600012888/YOPO-You-Only-Propagate-Once}}~\cite{zhang2019theoretically}, and Free\footnote{\url{https://github.com/ashafahi/free_adv_train}}~\cite{shafahi2019adversarial} with the hyper-parameters recommended in their works, and we select $1/\lambda$ as $6$ for TRADES (both ATTA and PGD).

In Section~\red{3}, which analyzes the transferability between training epochs, we use MAT with PGD-$10$ to train models and PGD-$20$ to calculate loss value and error rate.

Each experiment is taken on one idle NVIDIA GeForce RTX 2080 Ti GPU. Except PGD attack, we implement other attacks with Adversarial Robustness Toolbox~\cite{art2018}.

\section{Experiment details} \label{sec:exp-details}
\subsection{Qualitative study on training images}
\begin{figure}[ht]
  \centering
  \begin{subfigure}[b]{0.5\textwidth}
  \centering
  \includegraphics[height=3.5cm]{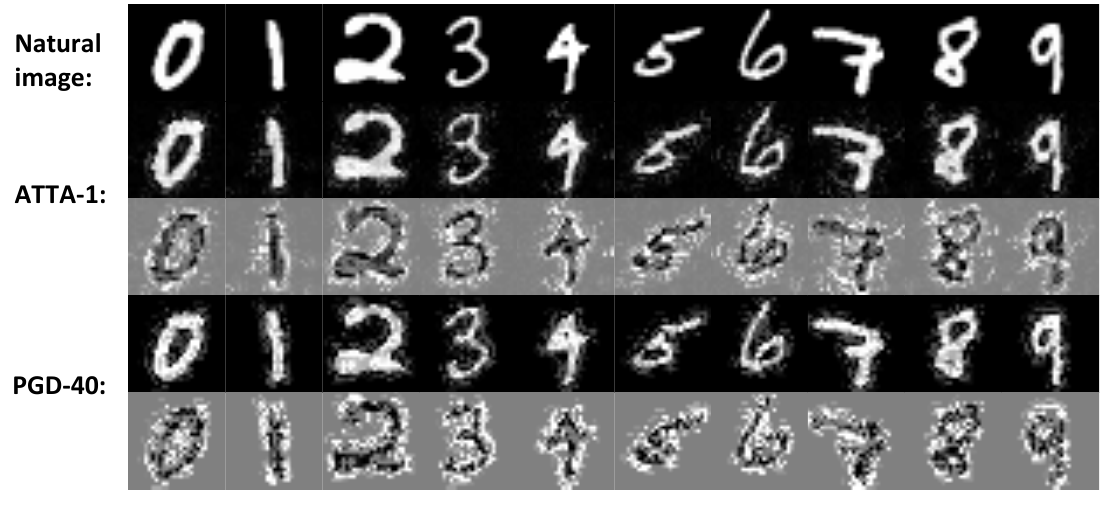}
  \caption{MNIST} \label{c:a}
  \end{subfigure}
  \begin{subfigure}[b]{0.5\textwidth}
  \centering
  \includegraphics[height=3.6cm]{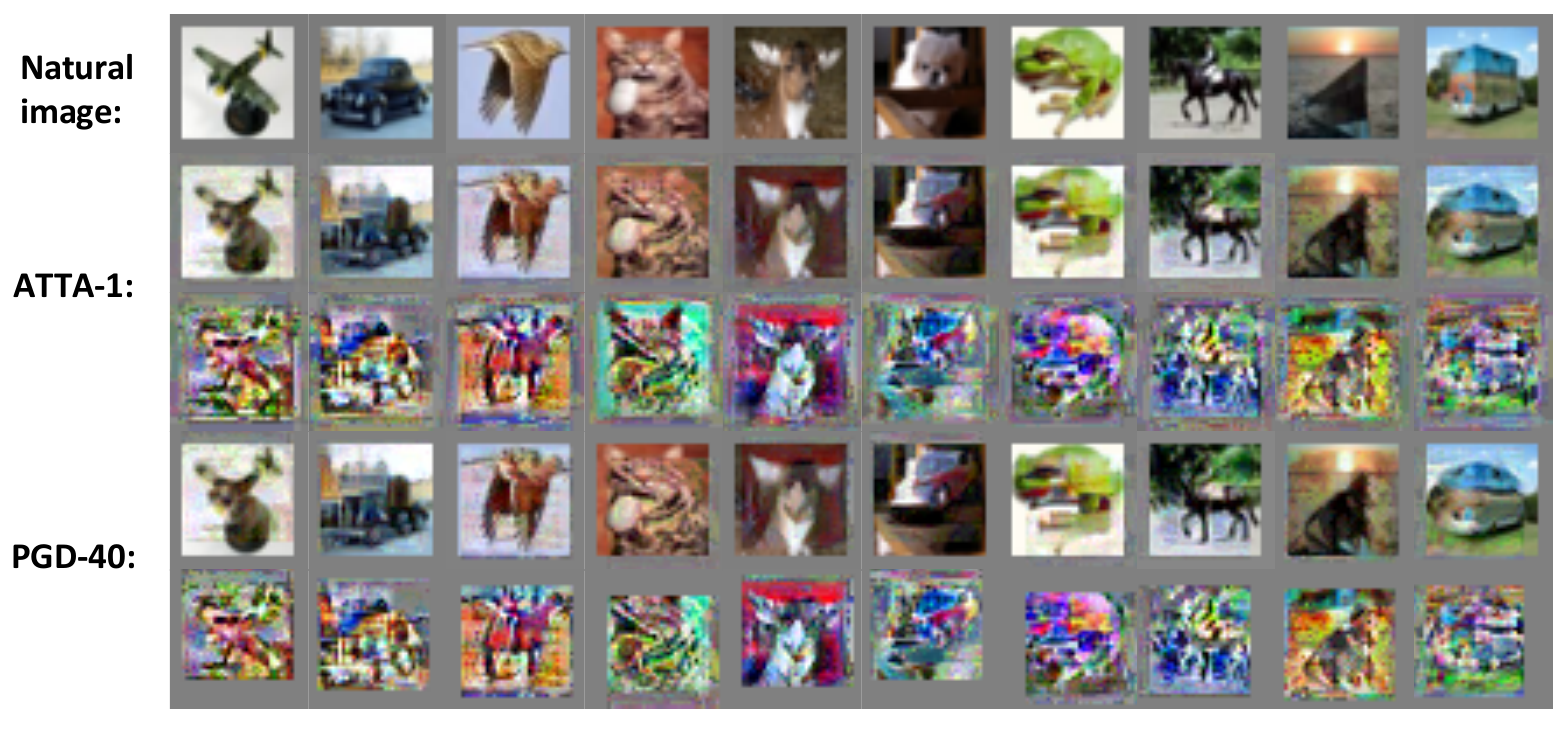}
  \caption{CIFAR10} \label{c:b}
  \end{subfigure}
  \caption{Visualization of natural images, adversarial examples and corresponding perturbations in each class for MNIST and CIFAR10. The first row in (a) and (b) shows the natural images. The second and third rows show the adversarial examples and perturbations generated by ATTA-$1$. The fourth and fifth rows show the adversarial examples and perturbations generated by PGD-$40$ in (a) and PGD-$10$ in (b).}
  \label{fig:case-study}
\end{figure}

To compare the quality of adversarial examples generated by PGD and ATTA, we visualize some adversarial examples generated by both methods. For MNIST, we choose the model checkpoint trained by MAT-ATTA-$1$ at epoch $40$. For CIFAR10, we choose the model checkpoint trained by MAT-ATTA-$1$ at epoch $30$.
Figure~\ref{fig:case-study} shows the adversarial examples and perturbations used to train the model (ATTA-$1$) and genereated by PGD-$40$ (PGD-$10$) attack on MNIST (CIFAR10) model in each class.
To better visualize the perturbation, we re-scale the perturbation by calculating $\frac{2x_p}{\epsilon}$ (where $x_p$ is the perturbation and $\epsilon$ is the $L_\infty$ bound of adversarial attack). This shifts the $L_\infty$ ball to the scale of $[0,1]$.

We find that, although ATTA-$1$ just performs one attack iteration in each epoch, it generates similar perturbations to PGD-$40$ (PGD-$10$) in MNIST (CIFAR10).
The effect of inverse data augmentation is shown in Figure~\ref{c:b}. There are some perturbations on the padded pixels in the third row (ATTA-$1$), but perturbations just generated by PGD-$10$ (shown in fifth row) just appear on cropped pixels.

\subsection{Complete evaluation results} \label{ssec:com-result}
We put the complete evaluation result in this section as a supplement to Section~\red{5.2}.

We evaluate defense methods under additional attacks and the evaluation results are shown in Table~\ref{tab:exp-mnist-ap} and Table~\ref{tab:exp-cifar-ap}.
Similar to the conclusion stated in Section~\red{5.2}, compared to other methods,
ATTA achieves comparable robustness with much less training time, which shows a better trade-off on training efficiency and robustness.
With the same number of attack iterations, ATTA needs less time to train the model. As mentioned in Section~\red{3.2}, with the accumulation of adversarial perturbations, ATTA can use the same number of attack iterations to achieve a higher attack strength, which helps the model converge faster.

Also, we test our models with stronger attacks which have a different attack size and multipule random starts. We evaluate ATTA-1 model trained with both MAT loss and TRADES loss against PGD-$20$ attack with an $\epsilon/4$ step-size and $20$ independent random starts.

\begin{table}[htp]
   \centering
   \begin{tabular}{cccc}
   \toprule
   \backslashbox{Defense}{Attack}& $\epsilon/4$ step-size & $20$ random starts\\ \midrule
   MAT-ATTA-1 &$49.56\%$& $48.48\%$  \\
   TRADES-ATTA-1 &$53.71\%$&$52.90\%$ \\ \bottomrule
   \end{tabular}
   \caption{Accuracy against PGD-$20$ attack with different settings.}\label{tab:abl-att-loss}
   \vspace{-0.3cm}
\end{table}

\section{Discussion}

\textbf{Natural accuracy v.s. Adversarial accuracy.} In this paper, we find that higher adversarial accuracy can lower natural accuracy.
This trade-off has been observed and explained in \cite{tsipras2018robustness, zhang2019theoretically}. A recent work~\cite{ilyas2019adversarial} points out that features used by naturally trained models are highly-predictive but not robust and adversarially trained models tend to use robust features rather than highly-predictive features, which may cause this trade-off.
Table~\ref{tab:exp-cifar-ap} also shows that, when models are trained with stronger attacks (more attack iterations), the models tend to have higher adversarial accuracy but lower natural accuracy.

\textbf{Transferability between training epochs.}  Adversarial attacks augment the training data to improve the model robustness. Our work points out that, unlike images augmented by traditional data augmentation methods that are independent between epochs, adversarial examples generated by adversarial attacks show high relevance transferability between epochs. We hope this finding can inspire other researchers to enhance adversarial training from a new perspective (e.g., improving transferability between epochs).

\begin{figure}[hpt]
   \centering
   \includegraphics[width=0.47\textwidth]{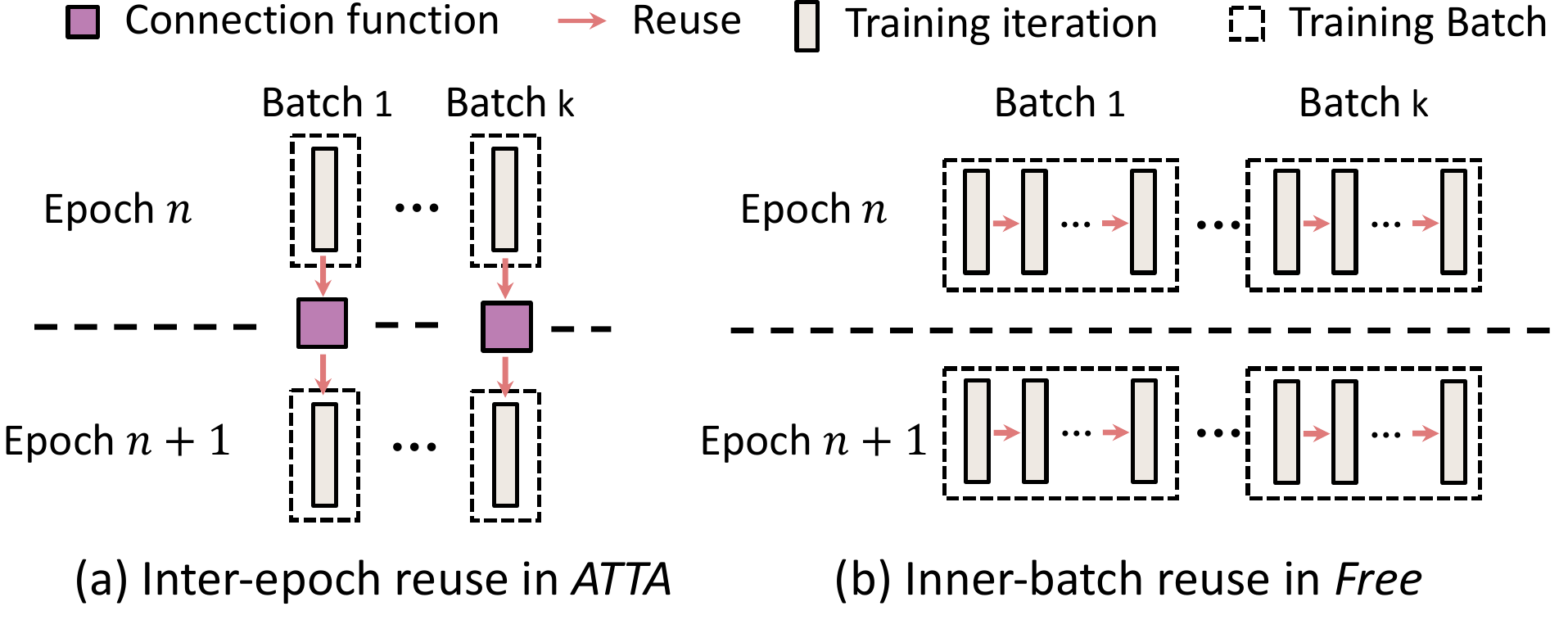}
   \caption{Comparison on reuse between Free and ATTA.}
   \label{fig:rebuttal}
\end{figure}

\textbf{Inter-epoch reuse in ATTA v.s. inner-batch reuse in Free.} Both ATTA and Free reuse perturbations during training. The difference between the two methods is how to reuse perturbations (Fig.~\ref{fig:rebuttal}). Free reuses perturbations through iterations on the top of the same batch of images (\textbf{inner-batch}),
but ATTA reuses perturbations across epochs (\textbf{inter-epoch}), which is complementary to inner-batch reuse.
Inter-epoch reuse is a non-trivial problem involving new challenges discussed in Section~4.1.
The notion of connection function, which is a crucial feature of ATTA to overcome these challenges, which only applies when exploiting transferability across epochs, does not exist in Free. We found that ATTA can outperform Free, even without any inner-batch reuse and Free's warm start between different batches (In Free, when a new batch comes, the perturbations of the previous batch are directly used as a warm start, but the mismatch between images and perturbations could be resulting in reduced effectiveness.).

\begin{table*}[h]
  \centering
  \begin{tabular}{c*{8}{c}}\toprule
  \multicolumn{2}{c}{\backslashbox{Defense}{Attack}}&Natural&PGD-40&PGD-100&M-PGD-40&FGSM&CW-20&\multicolumn{1}{c}{\begin{tabular}[c]{@{}c@{}}Time\\ (sec)\end{tabular}}\\\midrule
  \multirow{7}{*}{MAT}
  &PGD-1 & $99.52\%$&$15.82\%$&$4.17\%$&$13.99\%$&$81.84\%$&$98.99\%$&$226$\\
  &PGD-10 &$99.52\%$&$34.92\%$&$16.90\%$&$36.88\%$&$94.32\%$&$99.38\%$&$1649$\\
  &PGD-40 &$99.37\%$&$\textbf{96.21\%}$&$94.69\%$&$96.79\%$&$97.37\%$&$99.06\%$&$\textbf{3933}$\\\cmidrule{2-9}
  &YOPO-$5$-$10$ &$99.15\%$&$93.69\%$&$87.93\%$&$99.04\%$&$94.51\%$&$99.03\%$&$789$\\\cmidrule{2-9}
  &ATTA-1 &$99.45\%$&$\textbf{96.31\%}$&$95.11\%$&$96.15\%$&$98.31\%$&$99.14\%$&$\textbf{297}$\\
  &ATTA-10 &$99.41\%$&$97.36\%$&$96.75\%$&$97.45\%$&$98.37\%$&$99.3\%$&$1687$\\
  &ATTA-40 &$99.23\%$&$97.28\%$&$96.85\%$&$97.51\%$&$98.55\%$&$98.02\%$&$4650$\\
  \midrule\midrule

  \multirow{6}{*}{TRADES}
  &PGD-1 &$99.41\%$&$39.53\%$&$31.28\%$&$36.42\%$&$71.01\%$&$99.17\%$&$583$\\
  &PGD-10 &$99.37\%$&$50.06\%$&$25.71\%$&$50.59\%$&$95.12\%$&$99.24\%$&$1823$\\
  &PGD-40 &$98.89\%$&$\textbf{96.54\%}$&$95.54\%$&$96.79\%$&$97.83\%$&$98.87\%$&$\textbf{6544}$\\\cmidrule{2-9}
  &ATTA-1 &$99.03\%$&$\textbf{96.10\%}$&$94.24\%$&$95.86\%$&$98.38\%$&$98.93\%$&$\textbf{460}$\\
  &ATTA-10 &$98.83\%$&$96.86\%$&$96.34\%$&$96.90\%$&$98.11\%$&$98.68\%$&$1686$\\
  &ATTA-40 &$98.21\%$&$96.03\%$&$96.33\%$&$96.80\%$&$97.85\%$&$98.32\%$&$4660$\\
  \bottomrule
  \end{tabular}
  \caption{The result of various attacks on MNIST dataset.}
  \label{tab:exp-mnist-ap}
 \end{table*}

 \begin{table*}[h]
  \centering
  \begin{tabular}{c*{8}{c}}\toprule
  \multicolumn{2}{c}{\backslashbox{Defense}{Attack}}&Natural&PGD-20&PGD-100&M-PGD-20&FGSM&CW-20&\multicolumn{1}{c}{\begin{tabular}[c]{@{}c@{}}Time\\ (min)\end{tabular}}\\\midrule
  \multirow{10}{*}{MAT}
  &PGD-1 &$93.18\%$&$22.3\%$&$21.63\%$&$23.51\%$&$49.88\%$&$31\%$&$435$\\
  &PGD-2 &$91.33\%$&$35.16\%$&$34.46\%$&$38.61\%$&$55.84\%$&$44.12\%$&$600$\\
  &PGD-3 &$89.95\%$&$41.38\%$&$40.59\%$&$42.17\%$&$59.82\%$&$50.94\%$&$785$\\
  &PGD-10 & $87.49\%$& $\textbf{47.07\%}$&$46.77\%$&$47.95\%$&$63.5\%$&$56.7\%$&$\textbf{2027}$\\\cmidrule{2-9}
  &Free($m=8$) &$85.54\%$&$47.68\%$&$47.22\%$&$48.02\%$&$60.22\%$&$57.58\%$&$640$\\
  &YOPO-$5$-$3$ &$86.43\%$&$48.24\%$&$47.74\%$&$51.87\%$&$59.78\%$&$81.98\%$&$335$\\\cmidrule{2-9}
  &ATTA-1 &$85.71\%$&$\textbf{50.96\%}$&$48.18\%$&$52.31\%$&$64.16\%$&$76\%$&$\textbf{134}$\\
  &ATTA-2 &$85.95\%$&$51.90\%$&$49.45\%$&$53.08\%$&$64.42\%$&$77.02\%$&$196$\\
  &ATTA-3 &$85.44\%$&$52.56\%$&$50.77\%$&$53.92\%$&$64.82\%$&$75.92\%$&$267$\\
  &ATTA-10 & $83.80\%$& $54.33\%$&$52.6\%$&$55.38\%$&$63.49\%$&$75.37\%$&$690$\\
  \midrule\midrule

  \multirow{9}{*}{TRADES}
  &PGD-1 &$93.58\%$&$35.52\%$&$31.84\%$&$38.42\%$&$65.03\%$&$86.46\%$&$592$\\
  &PGD-2 &$90.61\%$&$49.75\%$&$45.66\%$&$51.19\%$&$60.86\%$&$85.48\%$&$730$\\
  &PGD-3 &$88.52\%$&$54.20\%$&$52.19\%$&$55.91\%$&$62.88\%$&$84.48\%$&$861$\\
  &PGD-10 &$84.13\%$&$\textbf{56.6}\%$&$55.36\%$&$57.76\%$&$63.02\%$&$79.4\%$&$\textbf{2028}$\\\cmidrule{2-9}
  &ATTA-1 &$85.04\%$&$54.50\%$&$52.66\%$&$55.70\%$&$64.04\%$&$82.62\%$&$199$\\
  &ATTA-2 &$84.58\%$&$54.94\%$&$53.27\%$&$56.17\%$&$63.72\%$&$81.92\%$&$261$\\
  &ATTA-3 &$84.23\%$&$\textbf{56.36\%}$&$54.85\%$&$57.24\%$&$64.03\%$&$81.82\%$&$\textbf{320}$\\
  &ATTA-10 &$83.67\%$&$57.34\%$&$56.39\%$&$58.15\%$&$64.1\%$&$82.01\%$&$752$\\
  \bottomrule
  \end{tabular}
  \caption{The result of various attacks on CIFAR10 dataset.}
  \label{tab:exp-cifar-ap}
 \end{table*}

%% file: egpaper_final.bbl
\begin{thebibliography}{10}\itemsep=-1pt

\bibitem{baluja2018learning}
Shumeet Baluja and Ian Fischer.
\newblock Learning to attack: Adversarial transformation networks.
\newblock In {\em AAAI}, pages 2687--2695, 2018.

\bibitem{cai2018curriculum}
Qi-Zhi Cai, Min Du, Chang Liu, and Dawn Song.
\newblock Curriculum adversarial training.
\newblock In {\em International Joint Conferences on Artificial Intelligence
  (IJCAI)}, 2018.

\bibitem{carlini2017adversarial}
Nicholas Carlini and David Wagner.
\newblock Adversarial examples are not easily detected: Bypassing ten detection
  methods.
\newblock In {\em Proceedings of the 10th ACM Workshop on Artificial
  Intelligence and Security}, pages 3--14. ACM, 2017.

\bibitem{carmon2019unlabeled}
Yair Carmon, Aditi Raghunathan, Ludwig Schmidt, Percy Liang, and John~C Duchi.
\newblock Unlabeled data improves adversarial robustness.
\newblock In {\em Advances in Neural Information Processing Systems}, 2019.

\bibitem{cohen2019certified}
Jeremy~M Cohen, Elan Rosenfeld, and J~Zico Kolter.
\newblock Certified adversarial robustness via randomized smoothing.
\newblock In {\em International Conference on Machine Learning (ICML)}, 2019.

\bibitem{imagenet_cvpr09}
J. Deng, W. Dong, R. Socher, L.-J. Li, K. Li, and L. Fei-Fei.
\newblock {ImageNet: A Large-Scale Hierarchical Image Database}.
\newblock In {\em Proceedings of the IEEE Conference on Computer Vision and
  Pattern Recognition (CVPR)}, 2009.

\bibitem{dong2018boosting}
Yinpeng Dong, Fangzhou Liao, Tianyu Pang, Hang Su, Jun Zhu, Xiaolin Hu, and
  Jianguo Li.
\newblock Boosting adversarial attacks with momentum.
\newblock In {\em Proceedings of the IEEE conference on computer vision and
  pattern recognition}, pages 9185--9193, 2018.

\bibitem{eykholt2017robust}
Kevin Eykholt, Ivan Evtimov, Earlence Fernandes, Bo Li, Amir Rahmati, Chaowei
  Xiao, Atul Prakash, Tadayoshi Kohno, and Dawn Song.
\newblock Robust physical-world attacks on deep learning models.
\newblock In {\em Proceedings of the IEEE Conference on Computer Vision and
  Pattern Recognition (CVPR)}, 2019.

\bibitem{goodfellow2014explaining}
Ian~J Goodfellow, Jonathon Shlens, and Christian Szegedy.
\newblock Explaining and harnessing adversarial examples.
\newblock In {\em International Conference on Learning Representations (ICLR)},
  2014.

\bibitem{he2016deep}
Kaiming He, Xiangyu Zhang, Shaoqing Ren, and Jian Sun.
\newblock Deep residual learning for image recognition.
\newblock In {\em Proceedings of the IEEE conference on computer vision and
  pattern recognition}, pages 770--778, 2016.

\bibitem{hein2017formal}
Matthias Hein and Maksym Andriushchenko.
\newblock Formal guarantees on the robustness of a classifier against
  adversarial manipulation.
\newblock In {\em Advances in Neural Information Processing Systems}, pages
  2266--2276, 2017.

\bibitem{hendrycks2019using}
Dan Hendrycks, Kimin Lee, and Mantas Mazeika.
\newblock Using pre-training can improve model robustness and uncertainty.
\newblock In {\em International Conference on Machine Learning (ICML)}, 2019.

\bibitem{ilyas2019adversarial}
Andrew Ilyas, Shibani Santurkar, Dimitris Tsipras, Logan Engstrom, Brandon
  Tran, and Aleksander Madry.
\newblock Adversarial examples are not bugs, they are features.
\newblock {\em arXiv preprint arXiv:1905.02175}, 2019.

\bibitem{jang2019adversarial}
Yunseok Jang, Tianchen Zhao, Seunghoon Hong, and Honglak Lee.
\newblock Adversarial defense via learning to generate diverse attacks.
\newblock In {\em Proceedings of the IEEE International Conference on Computer
  Vision}, pages 2740--2749, 2019.

\bibitem{krizhevsky2009learning}
Alex Krizhevsky, Geoffrey Hinton, et~al.
\newblock Learning multiple layers of features from tiny images.
\newblock Technical report, Citeseer, 2009.

\bibitem{kurakin2016adversarial}
Alexey Kurakin, Ian Goodfellow, and Samy Bengio.
\newblock Adversarial machine learning at scale.
\newblock In {\em International Conference on Learning Representations (ICLR)},
  2017.

\bibitem{lecun1998gradient}
Yann LeCun, L{\'e}on Bottou, Yoshua Bengio, Patrick Haffner, et~al.
\newblock Gradient-based learning applied to document recognition.
\newblock {\em Proceedings of the IEEE}, 86(11):2278--2324, 1998.

\bibitem{lee2017generative}
Hyeungill Lee, Sungyeob Han, and Jungwoo Lee.
\newblock Generative adversarial trainer: Defense to adversarial perturbations
  with gan.
\newblock {\em arXiv preprint arXiv:1705.03387}, 2017.

\bibitem{li2018learning}
Yingwei Li, Song Bai, Yuyin Zhou, Cihang Xie, Zhishuai Zhang, and Alan Yuille.
\newblock Learning transferable adversarial examples via ghost networks.
\newblock In {\em AAAI}, 2020.

\bibitem{liu2016delving}
Yanpei Liu, Xinyun Chen, Chang Liu, and Dawn Song.
\newblock Delving into transferable adversarial examples and black-box attacks.
\newblock In {\em International Conference on Learning Representations (ICLR)},
  2017.

\bibitem{madry2017towards}
Aleksander Madry, Aleksandar Makelov, Ludwig Schmidt, Dimitris Tsipras, and
  Adrian Vladu.
\newblock Towards deep learning models resistant to adversarial attacks.
\newblock In {\em International Conference on Learning Representations (ICLR)},
  2018.

\bibitem{art2018}
Maria-Irina Nicolae, Mathieu Sinn, Minh~Ngoc Tran, Beat Buesser, Ambrish Rawat,
  Martin Wistuba, Valentina Zantedeschi, Nathalie Baracaldo, Bryant Chen, Heiko
  Ludwig, Ian Molloy, and Ben Edwards.
\newblock Adversarial robustness toolbox v1.2.0.
\newblock {\em CoRR}, 1807.01069, 2018.

\bibitem{papernot2016transferability}
Nicolas Papernot, Patrick McDaniel, and Ian Goodfellow.
\newblock Transferability in machine learning: from phenomena to black-box
  attacks using adversarial samples.
\newblock {\em arXiv preprint arXiv:1605.07277}, 2016.

\bibitem{papernot2017practical}
Nicolas Papernot, Patrick McDaniel, Ian Goodfellow, Somesh Jha, Z~Berkay Celik,
  and Ananthram Swami.
\newblock Practical black-box attacks against machine learning.
\newblock In {\em Proceedings of the 2017 ACM on Asia conference on computer
  and communications security}, pages 506--519. ACM, 2017.

\bibitem{raghunathan2018certified}
Aditi Raghunathan, Jacob Steinhardt, and Percy Liang.
\newblock Certified defenses against adversarial examples.
\newblock In {\em International Conference on Learning Representations (ICLR)},
  2018.

\bibitem{schmidt2018adversarially}
Ludwig Schmidt, Shibani Santurkar, Dimitris Tsipras, Kunal Talwar, and
  Aleksander Madry.
\newblock Adversarially robust generalization requires more data.
\newblock In {\em Advances in Neural Information Processing Systems}, pages
  5014--5026, 2018.

\bibitem{shafahi2019adversarial}
Ali Shafahi, Mahyar Najibi, Amin Ghiasi, Zheng Xu, John Dickerson, Christoph
  Studer, Larry~S Davis, Gavin Taylor, and Tom Goldstein.
\newblock Adversarial training for free!
\newblock In {\em Neural Information Processing Systems (NeurIPS)}, 2019.

\bibitem{shafahi2018universal}
Ali Shafahi, Mahyar Najibi, Zheng Xu, John Dickerson, Larry~S Davis, and Tom
  Goldstein.
\newblock Universal adversarial training.
\newblock {\em arXiv preprint arXiv:1811.11304}, 2018.

\bibitem{sharif2016accessorize}
Mahmood Sharif, Sruti Bhagavatula, Lujo Bauer, and Michael~K Reiter.
\newblock Accessorize to a crime: Real and stealthy attacks on state-of-the-art
  face recognition.
\newblock In {\em Proceedings of the 2016 ACM SIGSAC Conference on Computer and
  Communications Security}, pages 1528--1540. ACM, 2016.

\bibitem{song2018physical}
Dawn Song, Kevin Eykholt, Ivan Evtimov, Earlence Fernandes, Bo Li, Amir
  Rahmati, Florian Tramer, Atul Prakash, and Tadayoshi Kohno.
\newblock Physical adversarial examples for object detectors.
\newblock In {\em 12th USENIX Workshop on Offensive Technologies (WOOT 18)},
  2018.

\bibitem{szegedy2013intriguing}
Christian Szegedy, Wojciech Zaremba, Ilya Sutskever, Joan Bruna, Dumitru Erhan,
  Ian Goodfellow, and Rob Fergus.
\newblock Intriguing properties of neural networks.
\newblock In {\em International Conference on Learning Representations (ICLR)},
  2014.

\bibitem{tramer2017ensemble}
Florian Tram{\`e}r, Alexey Kurakin, Nicolas Papernot, Ian Goodfellow, Dan
  Boneh, and Patrick McDaniel.
\newblock Ensemble adversarial training: Attacks and defenses.
\newblock In {\em International Conference on Learning Representations (ICLR)},
  2018.

\bibitem{tsipras2018robustness}
Dimitris Tsipras, Shibani Santurkar, Logan Engstrom, Alexander Turner, and
  Aleksander Madry.
\newblock Robustness may be at odds with accuracy.
\newblock In {\em International Conference on Learning Representations (ICLR)},
  2019.

\bibitem{wang2019bilateral}
Jianyu Wang and Haichao Zhang.
\newblock Bilateral adversarial training: Towards fast training of more robust
  models against adversarial attacks.
\newblock In {\em Proceedings of the IEEE International Conference on Computer
  Vision}, pages 6629--6638, 2019.

\bibitem{wong2017provable}
Eric Wong and J~Zico Kolter.
\newblock Provable defenses against adversarial examples via the convex outer
  adversarial polytope.
\newblock In {\em International Conference on Machine Learning (ICML)}, 2018.

\bibitem{zagoruyko2016wide}
Sergey Zagoruyko and Nikos Komodakis.
\newblock Wide residual networks.
\newblock {\em arXiv preprint arXiv:1605.07146}, 2016.

\bibitem{zhang2019you}
Dinghuai Zhang, Tianyuan Zhang, Yiping Lu, Zhanxing Zhu, and Bin Dong.
\newblock You only propagate once: Accelerating adversarial training via
  maximal principle.
\newblock In {\em Neural Information Processing Systems (NeurIPS)}, 2019.

\bibitem{zhang2019limitations}
Huan Zhang, Hongge Chen, Zhao Song, Duane Boning, Inderjit~S Dhillon, and
  Cho-Jui Hsieh.
\newblock The limitations of adversarial training and the blind-spot attack.
\newblock In {\em International Conference on Learning Representations (ICLR)},
  2019.

\bibitem{zhang2019theoretically}
Hongyang Zhang, Yaodong Yu, Jiantao Jiao, Eric~P Xing, Laurent~El Ghaoui, and
  Michael~I Jordan.
\newblock Theoretically principled trade-off between robustness and accuracy.
\newblock In {\em International Conference on Machine Learning (ICML)}, 2019.

\end{thebibliography}
